\documentclass{article}
\usepackage{ijcai25}

\usepackage{times}
\usepackage{soul}
\usepackage{url}
\usepackage[hidelinks]{hyperref}
\usepackage[utf8]{inputenc}
\usepackage[small]{caption}
\usepackage{graphicx}
\usepackage{amsmath}
\usepackage{amsthm}
\usepackage{booktabs}
\usepackage{algorithm}
\usepackage{algorithmic}
\usepackage[switch]{lineno}
\usepackage{color}
 \usepackage{multirow}
\usepackage{amssymb}

\title{MVP-CBM: Multi-layer Visual Preference-enhanced Concept Bottleneck Model \\ for Explainable Medical Image Classification}

\author{
Chunjiang Wang$^{1,2}$
\and
Kun Zhang$^{1,2,}$\thanks{Corresponding author} \and
Yandong Liu$^{1,2}$\and
Zhiyang He$^4$ \and
Xiaodong Tao$^4$ \\ \And
S. Kevin Zhou$^{1,2,3,5,}$\footnotemark[1]
\affiliations
$^1$School of Biomedical Engineering, Division of Life Sciences and Medicine,\\ University of Science and Technology of China, Hefei, Anhui, 230026, P.R. China\\
$^2$Suzhou Institute for Advanced Research, University of Science and Technology of China, \\Suzhou, Jiangsu, 215123, P.R. China\\
$^3$Jiangsu Provincial Key Laboratory of Multimodal Digital Twin Technology, \\Suzhou, Jiangsu, 215123, P.R. China\\
$^4$Medical Business Department, iFlytek Co., Ltd, Hefei, 230088, China\\
$^5$State Key Laboratory of Precision and Intelligent Chemistry, USTC
\emails
\{chunjiang\_wang, yandongliu\}@mail.ustc.edu.cn,  kkzhang@ustc.edu.cn, \\
\{zyhe, xdtao\}@iflytek.com,
skevinzhou@ustc.edu.cn
}

\begin{document}
\maketitle

\begin{abstract}

The concept bottleneck model (CBM), as a technique improving interpretability via linking predictions to human-understandable concepts, makes high-risk and life-critical medical image classification credible. Typically, existing CBM methods associate the final layer of visual encoders with concepts to explain the model’s predictions. 
However, we empirically discover the phenomenon of concept preference variation, that is, the concepts are preferably associated with the features at different layers than those only at the final layer; yet a blind last-layer-based association neglects such a preference variation and thus weakens the accurate correspondences between features and concepts, impairing model interpretability. To address this issue, we propose a novel Multi-layer Visual Preference-enhanced Concept Bottleneck Model (MVP-CBM), which comprises two key novel modules: (1) intra-layer concept preference modeling, which captures the preferred association of different concepts with features at various visual layers, and (2) multi-layer concept sparse activation fusion, which sparsely aggregates concept activations from multiple layers to enhance performance. Thus, by explicitly modeling concept preferences, MVP-CBM can comprehensively leverage multi-layer visual information to provide a more nuanced and accurate explanation of model decisions.
Extensive experiments on several public medical classification benchmarks demonstrate that MVP-CBM achieves state-of-the-art accuracy and interoperability, verifying its superiority. Code is available at \url{https://github.com/wcj6/MVP-CBM}.

\end{abstract}

\section{Introduction} \label{intro}

Deep neural networks have achieved remarkable success in various visual recognition tasks~\cite{he2016deep}, yet a key challenge they confront is the internal ``black box" decision-making process, lacking the desired model transparency and poses safety concerns especially for life-critical medical image classification~\cite{vuong2014molecular}, {\em e.g.}, cancer diagnosis.
Consequently, transforming medical image classification models into more interpretable ``white boxes” is both a critical and challenging endeavor~\cite{hou2024self}.


\begin{figure}[t]
\begin{center}
   \includegraphics[width=\linewidth]{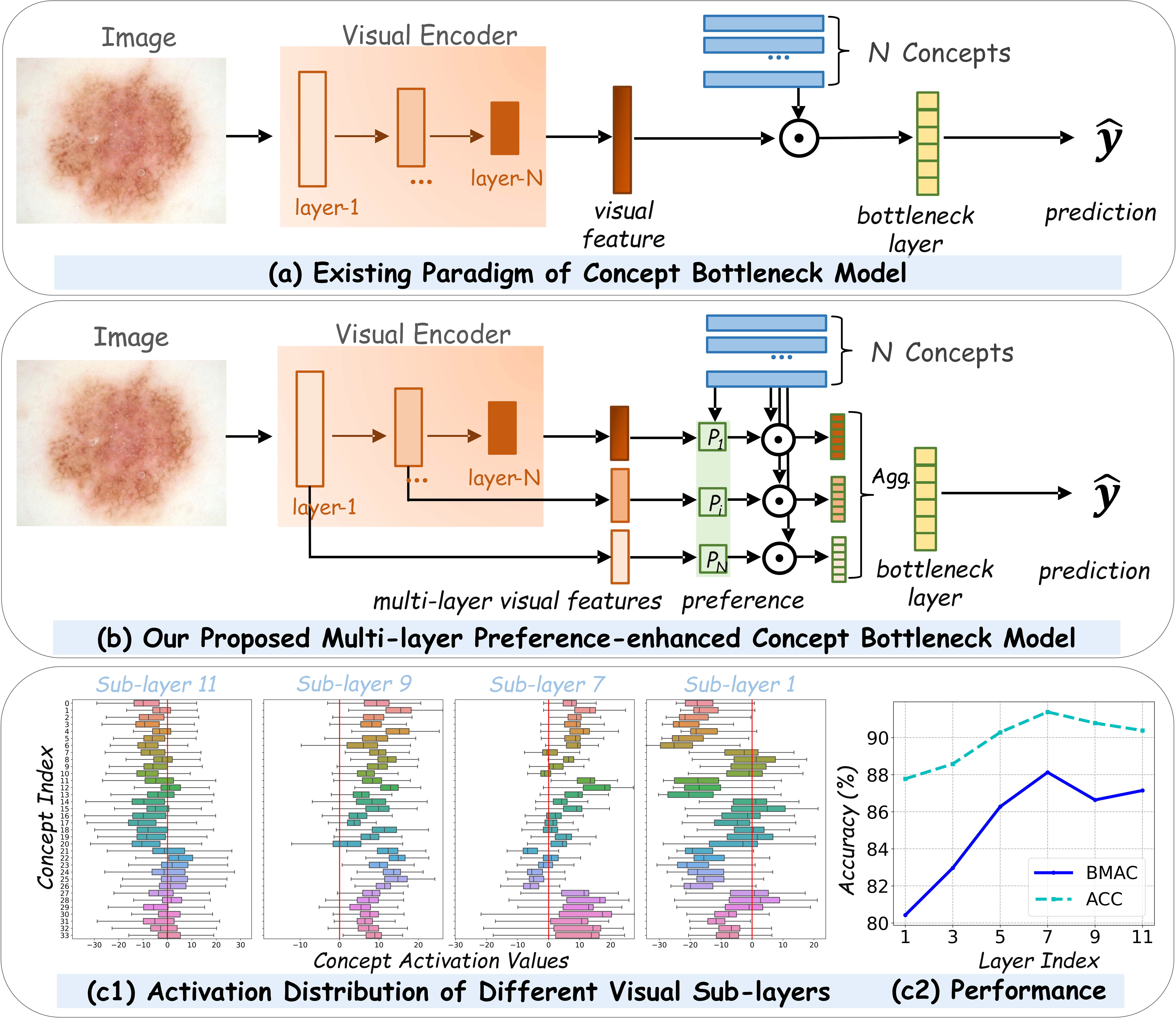}
\end{center}
   \caption{Illustration of our motivation. Unlike the existing CBM paradigm (Fig.a) that only uses features of the last standard layer of the visual encoder to correspond to concepts to explain the model’s prediction, we find that different visual sub-layer features exhibit varying preferences for concepts, i.e., some layer features are more inclined to activate and explain certain concepts, resulting in the default last standard layer may be not optimal (Fig.c). Thus, we propose to model the correspondence preference between visual sub-layer features and concepts, further comprehensively utilizing multi-layers to enhance the model’s interpretability and accuracy (Fig.b).}
\label{fig:motivation}
\end{figure}



In recent years, the Concept Bottleneck Model (CBM) has emerged as a promising interpretable neural network~\cite{koh2020concept}, providing explanations in terms of high-level, human-understandable concept attributes. Its key idea is to learn an intermediate bottleneck layer, in which each element corresponds to an interpretable concept, and constrain the model to make decisions based on this bottleneck layer. Therefore, the model's decision factors can be explicitly explained based on the concept activation values in the bottleneck layer.
The early CBM paradigm is based on bottleneck-layer supervised learning of manually annotated concepts~\cite{kim2023probabilistic,zarlenga2022concept}, which leads to labor-intensiveness and concept-incompleteness. 
Recent advancements in large language models have paved the way for automatically generated concepts~\cite{yang2023language,oikarinen2023label,shang2024incremental}, reducing the dependency on manual labeling and improving interpretability.
Generally, Fig.~\ref{fig:motivation}-(a) displays a common paradigm~\cite{gao2024aligning} that most existing CBM methods are based on. It first extracts the image features of the last standard layer of a pre-trained visual encoder, then calculates the similarities between the extracted visual features and concept textual features that are automatically constructed as the bottleneck layer activations, and finally utilizes the activations to make classification predictions.

However, we argue that the existing CBM paradigm typically ignores the so-called {\bf concept preference variation}, that is, visual features in different layers may exhibit varying association abilities to the concepts. 
Only leveraging the visual output of the last standard layer weakens the model's ability to accurately correspond to concepts, hence impairing interpretability.
As shown in Fig.~\ref{fig:motivation}-(c1), based on the existing state-of-the-art CBM-based skin disease diagnosis model~\cite{gao2024aligning}, we show the activation degree distribution of different visual layers and concepts, where each index on the vertical axis denotes a concept and the horizontal axis denotes the concept activation amplitude. For example, for the concept
of `1strawberry pattern, glomerular vessels' (index 18), the visual features of the 7th layer have a higher activation than those of the 9th and 11th layers, indicating that it prefers to recognize and explain this concept. It is generally understood that the shallow-to-deep layers in the visual extraction process are usually considered to encode different levels of information~\cite{yao2024deco}, which also leads to preference variance in correspondence with conceptual attributes~\cite{zhang2022negative}. 
Moreover, as depicted in Fig.~\ref{fig:motivation}-(c2), we evaluate the accuracy of concept prediction using layer-specific features, which further proves our motivation that the default standard output layer may not be optimal. As a result, ignoring these layer-specific preferences may hinder the accurate discovery of visual concepts and compromise the transparency of decision-making processes.

To address the above limitations, we propose a novel multi-layer visual preference-enhanced concept bottleneck model (MVP-CBM), which explicitly accounts for the concept preferences of each visual layer, and integrates concept activation fusion across layers to comprehensively enhance both the model’s interpretability and accuracy, as in Fig.~\ref{fig:motivation}-(b). 
Specifically, our method contains two modules: 1) Intra-layer Concept Preference Modeling (ICPM), which takes into account the preferences of different visual layers for different concepts. Rather than treating all layers equally, this approach recognizes that different layers may focus on different aspects of a concept, from low-level features to high-level semantic representations.
2) Multi-layer Concept Sparse Activation Fusion (MCSAF), which is designed to sparsely aggregate multi-layer concept activations to enhance model performance 
and interpretability. Instead of relying solely on the activations from the final output layer or a single concept bottleneck layer, MCSAF captures the concept-specific information at various intermediate layers, each of which may focus on different aspects of visual or semantic features.

Our contributions are summarized as follows:
\begin{itemize}

\item To the best of our knowledge, we are the first to discover the concept preference variance across visual layers in the CBM and leverage this to improve model interpretability.

\item We propose a novel method, MVP-CBM, which captures the preferences of different concepts across various visual layers and sparsely aggregates multi-layer concept activations to enhance model performance.

\item Through extensive experiments on several public medical classification benchmarks, MVP-CBM achieves state-of-the-art accuracy and interpretability, outperforming existing CBM methods and black-box models.
\end{itemize}



\begin{figure*}
  \centering
  \includegraphics[width=1.0\textwidth]{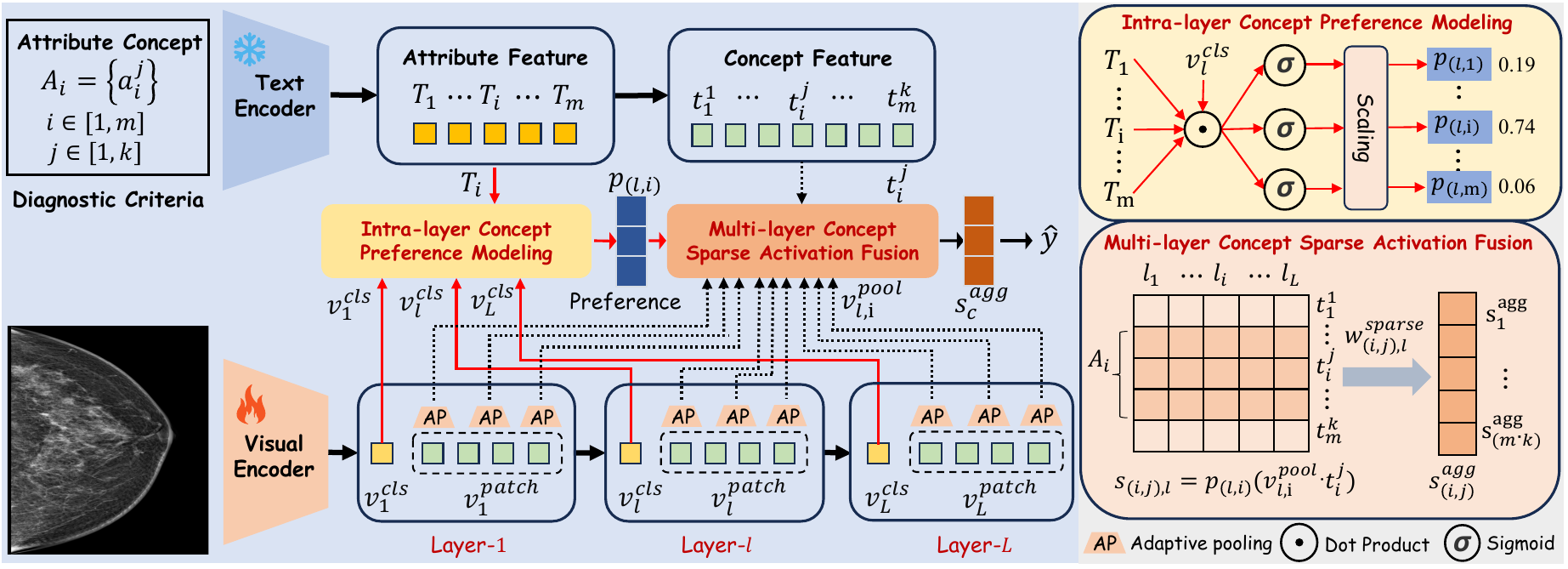}
  \caption{MVP-CBM consists of two key components: 1) Intra-layer Concept Preference Modeling (ICPM) and 2) Multi-layer Concept Sparse Activation Fusion (MCSAF). 1) In ICPM, intra-layer concept preferences are captured across different layers using the attribute global-level feature and the class token of each visual layer, followed by scaling. 2) In MCSAF, fine-grained concept activation scores for each concept are computed using the concept's local-level feature and the pooled patch token from each visual layer, weighted by inter-layer preference. These concept activations from multiple layers are then sparsely aggregated to form a concept bottleneck layer, thereby enhancing both the model's performance and interpretability.
  }
  \label{figs:hfvca}
\end{figure*}

\section{Related Work}

Deep neural networks have shown exceptional performance in medical image classification, but their black-box nature limits trust among healthcare professionals and patients~\cite{hou2024self}. To improve interpretability and reliability, Concept Bottleneck Models (CBMs) have been developed to map complex visual features to human-understandable concepts, providing explanations for model decisions.

Recent advancements in CBMs have primarily focused on addressing the challenge of manual concept annotation. For example, PCBM~\cite{yuksekgonul2022post} leverages ConceptNet~\cite{speer2017conceptnet} to generate a concept library and calculates the projection distance between CLIP text concept embeddings and visual representations to define concept bottlenecks, reducing the need for manual labeling. Building upon this, LaBo~\cite{yang2023language} enhances CBMs by employing large language models (LLMs) to generate candidate concepts and utilizing sub-model optimization blocks to select the most relevant ones. Explicit~\cite{gao2024aligning} further advances this line of work by leveraging LLMs and Q-former mechanisms to improve automation and flexibility in medical concept alignment~\cite{zhang2022unified}.

Despite these advancements, existing CBM methods primarily match visual representation with concept embeddings at the standard output layer, overlooking the fact that different concepts may prefer different visual encoder layers, which can compromise decision-making transparency. Our work addresses this limitation by introducing two key innovations: (1) constructing concept preference for each visual sub-layer, which ensures that each concept is aligned with the most relevant features of the model, and (2) applying multi-layer concept sparse fusion, which aggregates concept activations from multiple layers to improve transparency and interpretability.

\section{Methodology}
\subsection{Overview of MVP-CBM}

The overall framework of our proposed Multi-Layer Visual Preference Concept Bottleneck Model (MVP-CBM) is shown in Fig.~\ref{figs:hfvca}, which aims to incorporate human-understandable concepts into the decision-making process, thus making the `black box' model more transparent and comprehensible for humans to increase trust in the model's predictions. 
Formally, the input data is denoted as $\mathcal{D} = \{(x, y)\}$, where $ x $ represents the image and $ y\in \mathcal{Y} $ is its corresponding disease category. The disease $ y $ is generally defined with a set of diagnostic attributes $ \mathcal{A} = \{A_i\}_{i=1}^m $, where each attribute $ A_i=\{a^j_i\}_{j=1}^k $ comprises $ k $ distinct concepts, which can be generated by large language models (e.g., ChatGPT) or human experts.

As we discussed in the Sec.~\ref{intro}, existing CBM methods typically predict based on the bottleneck layer activations, i.e., calculating between the image feature $v$ extracted on the last standard layer of a pre-trained visual encoder and the predefined concepts as:
\begin{equation}
    \hat{y} = \Psi(a^{1}_{1}v^T,\dots,a^{j}_{i}v^T,\dots, a^{k}_{m}v^T),
    \label{eq:1}
\end{equation}
where $ \Psi$ is a linear layer to perform label prediction.

However, our preliminary investigations, as illustrated in Fig.~\ref{fig:motivation} (c1) and (c2), reveal that different visual layers exhibit distinct preferences for various aspects of concepts. These preferences range from low-level features to high-level semantic representations. Relying solely on the visual features of the standard output layer may result in inconsistencies between concepts and visual data, potentially compromising the fidelity and interpretability of the model. Therefore, we propose a novel framework that first models the preferences of each layer for different concepts and then integrates concept activations across multiple layers. Firstly,  we define the concept preference function for a given layer $ \ell $ as follows:
\begin{equation}
    p_{\ell, i} = f_{preference}(v_{\ell}, A_{i}),
    \label{eq:2}
\end{equation}
where $p_{\ell, i}$ denotes the captured preference degree between the $\ell$-th visual layer's feature $v_l$ and the $i$-th concept attribute.

Then, the multi-layer preference-enhanced concept bottleneck layer activations are calculated to make prediction as:
\begin{equation}
    \hat{y} = \Psi\left( \text{agg}\left( \left\{ p_{\ell,i} \cdot a_i^j v_\ell^{T} \right\}_{\ell,i} \right) \right),
    \label{eq:3}
\end{equation}
in which $\ell \in [1, L], i \in [1,m]$, and $L$ denotes the total number of visual layers, and $\text{agg}(\cdot)$ represents the fusion of concept activations across all layers. This design ensures that contributions from all layers are appropriately and comprehensively considered, leading to more accurate and interpretable concept bottleneck model.





\subsection{Key Components of MVP-CBM}

\subsubsection{Visual Representation}

Different from previous CBM methods only using the last standard layer of the pre-trained visual encoder, we extract each visual layer feature of the encoder as:
\begin{equation}
    [v_\ell^{cls}, \mathbf{V}_\ell^{patch}] = \Phi_{image}(x),
\end{equation}
where the class token $v_\ell^{cls} \in \mathbb{R}^d $ and patch tokens $\mathbf{V}_\ell^{patch} = \{v_{\ell,1}^{patch}, v_{\ell,2}^{patch}, \dots, v_{\ell,N_p}^{patch}\} \in \mathbb{R}^{N_p \times d}$ denote global and local features, respectively, in a specific ViT model layer \( \ell \in [1, L] \). \( \mathbf{v}_{\ell,n}^{patch} \in \mathbb{R}^d \) denote the $n$-th patch token from layer \( \ell \), and \( N_p \) is the total number of patches. Same with the existing method~\cite{gao2024aligning}, the visual encoder is adopted as the BioMedCLIP model~\cite{zhang2023biomedclip}.

\begin{figure}[t]
    \centering
    \includegraphics[width=\linewidth]{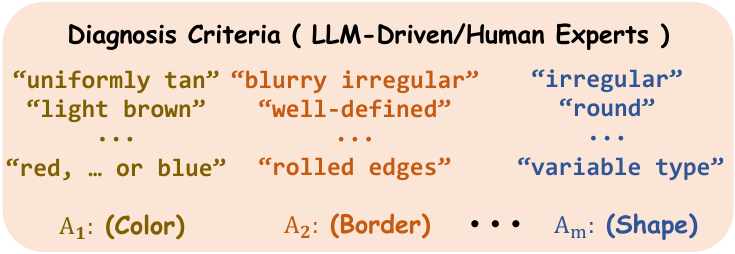}
    \caption{Concept examples for ISIC-2018 Dataset.}
    \label{fig:concept}
\end{figure}

\subsubsection{Concept Representation}

As shown in Fig.~\ref{fig:concept}, to define concept representation in medical imaging diagnosis, we start with identifying key disease-related concepts that are crucial for diagnosing specific conditions. These concepts are derived from medical literature, clinical guidelines, and expert knowledge~\cite{gao2024aligning}, ensuring they capture important diagnostic criteria. Specifically, the pre-trained BiomedCLIP's text encoder is used to extract the representation of each concept $a^j_{i}$ as:
\begin{equation}
     t^j_{i} = \Phi_{text}(a^j_{i}), \ i\in [1,m], \ j\in[1,k],
\end{equation}
where $t^j_{i} \in \mathbb{R}^{d}$ represents concept-based feature. $m$ is the number of diagnostic attributes, $k$ is the number of concepts per attribute, and $d$ refers to the dimension of text features. 


\subsubsection{Intra-layer Concept Preference Modeling (ICPM)}
%


Here, in contrast to existing methods that ignore the variance of the corresponding preferences between different visual layers and concepts~\cite{gao2024aligning}, our goal is to explicitly reveal the degree of concept preference at each visual layer. 
Considering that the concepts within an attribute reflect the same characteristic, we propose to quantify the preference value through the semantic correspondence between the global concept attribute feature and the visual layer features, {\em i.e.}, the degree to which the visual layer tends to explain the concept attribute.

Specifically, we concatenate all concepts within the same attribute to form an input that allows us to obtain an attribute global-level feature \( T_{i} \) for each $ A_i=\{a^j_i\}_{j=1}^k $. Formally, 
\begin{equation}
T_{i} = \Phi_{text}(\text{Concat}(a^1_{i}, \dots, a^j_{i}, \dots, a^k_{i})),
\end{equation}
where \( T_{i} \in \mathbb{R}^{d} \) encapsulates the collective semantic information of all concepts under a diagnostic attribute.

Then, in the intra-each visual layer, we compute the semantic correspondences between different attribute features \( \{T_{i}\}_{i=1}^{m} \) and the visual layer feature \( v_\ell^{cls} \) as: 

\begin{equation}
    p_{\ell,i} =\sigma \left( \frac{ v_\ell^{cls}T_{i}}{\|v_\ell^{cls}\| \|T_{i}\|} \right), \ i\in [1, m],
\end{equation}
where $\sigma(\cdot)$ denotes the Sigmoid activation function. 

The value of \(  p_{\ell, i} \) reflects the probability of association ability between the $\ell$-th visual layer and different attributes, in which the larger \(  p_{\ell, i} \) is, the more helpful the visual layer is in preferentially explaining the $i$-th concept attribute. 
Moreover, to further amplify the preference degree between different attributes, a learnable scaling parameter  \(\tau_1 \) is introduced. Therefore, the preference degree is calculated as follows:

\begin{equation}
    p_{\ell,i} = \frac{e^{p_{\ell,i}/\tau_1}}{\sum_{i=1}^{m} e^{p_{\ell,i}/\tau_1}}\in \mathbb{R}^{1 \times m}, \quad i \in [1,m],
\end{equation}
where $p_ {\ell, i}$ is designed to emphasize the preference for different attributes within the $\ell$-th layer.

\subsubsection{Multi-layer Concept Sparse Activation Fusion (MCSAF)}
To fully exploit inter-layer concept preferences while reducing redundancy introduced by inter-layer concept activations, we propose a multi-layer concept sparse activation fusion method. First, we compute fine-grained concept activation scores, weighted by the visual layer preferences. Next, to enhance the effectiveness of the most relevant visual layers for each concept, we calculate sparse activation weights for different visual layers. Finally, we aggregate the most representative concept activation scores after applying sparse activation to make reliable decisions.

Firstly, to alleviate the issue of an excessive number of local visual patch tokens in each visual layer, for the $i$-th diagnostic attribute, we only aggregate fine-grained semantic features related to the corresponding attribute from different layers, which is calculated as:

\begin{equation}
    v_{\ell,i}^{pool}=f_{\ell, i}(\mathbf{V}_\ell^{patch}), \ \ell \in [1,L], 
\end{equation}
in which  $f_{\ell, i}$ denotes the adaptive average pooling that down-samples the local visual patch tokens. $v_{\ell,i}^{pool} \in \mathbb{R}^{d}$ represents the pooled semantic feature corresponding to the $i$-th diagnostic attribute in the $\ell$-th visual layer.

The activation score of the concept within the $i$-th attribute across all layers is expressed as: 
\begin{equation} s_{(i,j),\ell} = p_{\ell,i} \cdot \left( \frac{ v_{\ell,i}^{pool}t^j_{i} }{\|v_{\ell,i}^{pool}\| \|t^j_{i}\|} \right), \ \ell \in [1, L], \end{equation} 
where the attribute preference $p_{\ell,i}$ is exploited to weight the concept activation scores. This ensures that each concept activation score is adjusted according to the corresponding attribute preference degree. That is, the concept activation, whose attribute has a higher association preference, is amplified, and vice versa.

Afterward, since different attributes exhibit varying preferences for visual layers, and not all concepts within a layer contribute equally to the final prediction, we propose a method to sparsify the less relevant visual layers. This maximizes the utilization of the most effective layers while reducing potential redundancy, improving the model's robustness. Concretely, we compute the concept contribution weights across all layers as follows:

\begin{equation}
    w_{(i,j),\ell} = \frac{e^{s_{(i,j),\ell}}}{\sum_{\ell=1}^L e^{s_{(i,j), \ell}}},
\end{equation}

To sparsify these weights, we introduce a thresholding mechanism that adaptively adjusts the activation weights of concepts across different visual layers.

\begin{equation} w^{adjust}_{(i,j),\ell} = e^{\tau_2(|w{(i,j),\ell}| - \theta_{\ell})} \odot w_{(i,j),\ell}, \end{equation}
where $e^{\tau_2(\cdot)}$ serves as an adjustment factor to scale $w_{(i,j),\ell}$ to introduce sparsity. Here, $\tau_2$ is a learnable scaling parameter. The threshold $\theta_{\ell}$ is adaptively adjusted based on the importance of each concept’s activation weights, as defined by: $\theta_{\ell} = \sigma(K)(w{_{\ell}}_{\text{max}} - w{_{\ell}}_{\text{min}}) + w{_{\ell}}_{\text{min}}$,
where \( K \) is a learnable parameter that modulates the threshold within the range of \([ w{_{\ell}}_{\text{min}}, w{_{\ell}}_{\text{max}} ]\).

Furthermore, we define a binary hard mask \( \mathbf{M}_{(i,j),\ell} \) for each layer \( \ell \) to filter out insignificant activation weights:
\begin{equation}
    \mathbf{M}_{(i,j),\ell} = 
    \begin{cases} 
        1 & \text{if } |w_{(i,j),\ell}| \geq \theta_\ell, \\
        0 & \text{otherwise}.
    \end{cases}
\end{equation}
Then, the sparse activation weights are obtained by applying this binary mask:
\begin{equation}
    w_{(i,j),\ell}^{\text{sparse}} =\mathbf{M}_{(i,j),\ell} \odot w^{adjust}_{(i,j),\ell},
\end{equation}
where activation weights with absolute contributions below the threshold \( \theta_{\ell} \) are compressed towards zero to effectively avoid their interference.

Finally, we aggregate the concept activation scores for each concept across all layers:
\begin{equation}
    s_{(i,j)}^{\text{agg}} = \sum_{\ell=1}^L (w_{(i,j),\ell}^{\text{sparse}} \odot s_{(i,j),\ell}),
\end{equation}
where $s_{(i,j)}^{\text{agg}}$ denotes the fused concept activation, which comprehensively and sparsely combines the contributions from multi-layers.

To encourage sparsity in the activation weights, we introduce a sparse loss term defined as the mean of the binary masks across all layers: 
\begin{equation}
    \mathcal{L}_{\text{sparse}} = \frac{1}{L} \sum_{\ell=1}^L \frac{1}{m \times k} \sum_{i=1}^m \sum_{j=1}^k \mathbf{M}_{(i,j),\ell},
\end{equation}
which penalizes the average number of active concepts, thereby promoting fewer active concepts and enhancing the model's interpretability.

\subsubsection{Classification and Training}


The fused concept activations of all concepts $s_{(i,j)}^{\text{agg}}$ are concatenated into a vector \( s_c^{\text{agg}} \in \mathbb{R}^{1 \times (m\cdot k)} \), facilitating subsequent classification layer to obtain predicted disease logits \( \hat{y} \):

\begin{equation}
    \hat{y} = \Psi(s_c^{\text{agg}})  \in \mathbb{R}^{|\mathcal{Y}|},
\end{equation}
where $ \Psi: \mathbb{R}^d \rightarrow \mathcal{Y} $ denotes a linear classifier.


The model is trained end-to-end by minimizing the combined loss:

\begin{equation}
    \mathcal{L} = \mathcal{L}_{ce}(y,\hat{y})+\lambda_1\mathcal{L}_{ce}(y_c,s_c)+\lambda_2\mathcal{L}_{\text{sparse}},
\end{equation}
where $y$ and $y_c$ are the disease label and concept label, respectively. $s_c \in  s_c^{\text{agg}}$ is the specfic concept activation score. \( \lambda_1\) and \(\lambda_2 \) are hyperparameters that balance the importance of accurate concept prediction and sparsity in concept activations. 
By jointly optimizing these parameters, the model learns to identify and emphasize the most relevant concepts across layers while maintaining sparsity, thereby enhancing both interpretability and predictive performance.

\section{Evaluation}

\begin{table*}[tbp]
\centering
\resizebox{\linewidth}{!}{%
\begin{tabular}{l|l|c|c|c|c|c|c|c}
\hline
\multirow{1}{*}{Setting} & \multirow{1}{*}{Model} & \multicolumn{1}{c|}{ISIC2018}& \multicolumn{1}{c|}{NCT} & \multicolumn{1}{c|}{IDRiD} & \multicolumn{1}{c|}{BUSI} & \multicolumn{1}{c|}{CMMD} & \multicolumn{1}{c|}{CM} & \multicolumn{1}{c}{SIIM} \\ 

\hline
Zero-shot & CLIP~\cite{radford2021learning} & 21.32 & 26.71 & 25.92 & 37.85 & 34.00 & 49.12 &  22.15 \\
          & MedCLIP~\cite{wang2022medclip} & 15.31 & 12.05 & 19.57 & 43.19 & 30.72 & 46.66 & 41.01 \\
          & BiomedCLIP~\cite{zhang2023biomedclip} & 24.47 & 40.39  & 30.31 & 36.45 & 50.00 & 46.70 &  20.92 \\
\hline
Black-box & ResNet50~\cite{he2016deep} & 80.25±0.80 & 92.34±0.64 & 56.67±1.95 & 84.39±1.36 & 69.51±2.79 & 79.21±1.50 & 81.01±0.74\\
          & ViT Base~\cite{dosovitskiy2020image}  & 87.31±1.17 &  93.33±0.69 & \textbf{66.61±0.81} & 85.87±0.58 &  68.12±0.79 & 80.30±1.30 & 80.48±1.04
          \\
\hline  
Explainable 
    & LaBo~\cite{yang2023language} & 82.63±0.40 & 90.12±0.74 & 52.77±0.78 & 84.08±0.80 & 64.49±0.82 & 72.53±0.60 & 73.46±0.69 \\
    
    &  PCBM~\cite{yuksekgonul2022post}  & 84.46±0.78  & 92.78±0.91 &  57.15±1.02   & 87.71±0.82  & 65.18±0.63  & 77.81±0.85 & 76.78±0.71 \\
            & Explicd~\cite{gao2024aligning} & 86.85±0.94 & 94.73±0.72 & 64.02±0.73 & 90.45±0.97 & 67.12±0.94 & 81.68±0.64 &  81.96±0.93\\
            
            & \textbf{MVP-CBM (Ours)} & \textbf{87.83±1.17} & \textbf{97.84±0.98}& 65.50±0.49 & \textbf{92.20±0.84} & \textbf{74.87±0.88} & \textbf{82.77±0.48} & \textbf{83.62±0.69} \\
\hline 
\end{tabular}%
}
\caption{Balance accuracy(BMAC)\%  is reported for all datasets. The best results are highlighted in bold.}
\label{tab:bmac}
\end{table*}

\begin{table*}[tbp]
\centering
\resizebox{\linewidth}{!}{%
\begin{tabular}{l|l|c|c|c|c|c|c|c}
\hline
\multirow{1}{*}{Setting} & \multirow{1}{*}{Model} & \multicolumn{1}{c|}{ISIC2018}& \multicolumn{1}{c|}{NCT} & \multicolumn{1}{c|}{IDRiD} & \multicolumn{1}{c|}{BUSI} & \multicolumn{1}{c|}{CMMD} & \multicolumn{1}{c|}{CM} & \multicolumn{1}{c}{SIIM} \\ 
\hline
Zero-shot & CLIP~\cite{radford2021learning} & 29.88 & 26.67 & 29.84 & 43.85 & 27.10 & 54.81 & 41.01 \\
          & MedCLIP~\cite{wang2022medclip} & 13.68 & 9.57 & 16.86 & 27.44 & 27.10 & 45.94 & 42.34 \\
          & BiomedCLIP~\cite{zhang2023biomedclip} & 59.83 &  40.51 & 39.15 & 29.36 & 52.90 & 49.13 & 39.88 \\
\hline
Black-box & ResNet50~\cite{he2016deep} & 82.23±0.86 &  93.40±0.87 & 57.82±2.25 & 78.79±1.50 & 63.47±2.14 & 79.98±1.01 & 81.79±0.93 \\
          & ViT Base~\cite{dosovitskiy2020image} & 89.34±0.97& 94.40±0.76 & 59.86±1.76 & 80.09±1.51 & 66.31±0.97 & 79.74±1.10 & 81.28±0.78 \\
\hline  
Explainable 
    & LaBo~\cite{yang2023language}  & 80.70±0.64 & 90.43±0.48 & 48.97±0.59 & 82.91±0.48 & 68.88±0.69 & 74.08±0.59 & 75.33±0.71  \\
    
    &  PCBM~\cite{yuksekgonul2022post}  & 84.71±0.69  & 93.27±0.67 & 52.46±0.94 &  86.04±0.76  & 67.94±0.80 & 78.88±0.78  &  78.92±0.90 \\

            & Explicd~\cite{gao2024aligning}  & 90.44±1.21 & 95.29±0.53& 58.63±0.86 & 89.89±0.68 & 72.31±0.79 & 77.87±0.50 & 84.28±0.63 \\
            
            & \textbf{MVP-CBM (Ours)} & \textbf{91.04±1.26} & \textbf{98.17±0.88}  & \textbf{61.66±0.67} & \textbf{93.29±1.04} & \textbf{74.01±0.73} &\textbf{80.78±0.51} & \textbf{85.61±0.73}  \\
\hline
\end{tabular}%
}
\caption{Accuracy(ACC)\%  is reported for all datasets. The best results are highlighted in bold.}
\label{tab:acc}
\end{table*}


\textbf{Datasets}: (1) \textbf{ISIC2018}~\cite{tschandl2018ham10000} contains 10015 dermoscopic images with seven skin lesion categories for skin cancer classification. (2) \textbf{NCT-CRC-HE}(NCT)~\cite{kather2018100} includes
100k patch-based histological images of human cancer with nine tissue classes for classification. (3)  \textbf{IDRiD}~\cite{porwal2018indian} consists of 455 retinal fundus images annotated with 5 severity level grading of diabetic retinopathy. (4) \textbf{BUSI}~\cite{al2019deep} dataset contains 780 ultrasound images of breast masses categorized into normal, benign, and malignant classes for breast cancer classification. (5) \textbf{CMMD}~\cite{cui2021chinese} includes 1,775 x-ray images with benign and malignant.  (6) \textbf{Cardiomegaly} (CM)~\cite{johnson2019mimic} is used for binary classification, which is the x-ray image in IMIC-CXR. (7) \textbf{SIIM-ACR Pneumothorax}(SIIM)~\cite{siim-acr-pneumothorax-segmentation} has 12,047 x-ray images for binary classifcation.

\textbf{Implementation Details}: We prompt GPT-o1 to query domain knowledge and diagnostic criteria. The details are listed in appendix. MVP-CBM and Explicd are implemented based on BioMedCLIP for all datasets. The fine-tuning of MVP-CBM and Explicd involves optimizing visual encoder, visual concept learning module, and the final linear layer with AdamW optimizer, while keeping the text encoder fixed. For the hyperparameters in MVP-CBM, the initial values are set as follows: $\tau_1$ and $\tau_2$ are both initialized to 0.2, and $K$ is initialized to 0. All experiments are conducted using PyTorch with an Nvidia 4090 GPU. To reduce randomness, we run 5 times and report mean and standard deviation. Given class imbalance in medical data, we report both balanced accuracy (BMAC) and accuracy (ACC) for comprehensive evaluation.


\subsection{Comparison with State-of-the-arts}

We compare our method against several baselines: (1) Zero-shot Models, including CLIP and specialized biomedical VLMs MedCLIP and BiomedCLIP~\cite{wang2022medclip,zhang2023biomedclip}, which perform classification without task-specific fine-tuning; (2) Supervised Black-box Models, such as ResNet50~\cite{he2016deep} and Vision Transformer(ViT)-based models~\cite{dosovitskiy2020image}, both fine-tuned for the classification task; and (3) Explainable Models, including LaBo~\cite{yang2023language}, PCBM~\cite{yuksekgonul2022post}, and the state-of-the-art Explicd~\cite{gao2024aligning}, which utilizes predefined diagnostic criteria for interpretable medical classification.

As shown in Table~\ref{tab:bmac} and Table~\ref{tab:acc}, Vision-Language Models (VLMs) like CLIP, MedCLIP, and BiomedCLIP perform poorly in zero-shot settings, with CLIP near random levels. Although BiomedCLIP achieves higher accuracy on the CMMD dataset due to its medical pretraining, it still lags behind supervised models such as ResNet50 and ViT-Base on other datasets, highlighting limited generalization.

In the explainable setting, LaBo shows lower accuracy because it relies on well-aligned VLMs, highlighting the challenge of balancing accuracy and explainability. PCBM improves upon LaBo by utilizing a robust concept bank but still falls short of Explicd, which integrates human knowledge and visual concept learning. In contrast, MVP-CBM integrates concept preferences from multiple layers of the visual encoder, aligning diverse visual features with diagnostic criteria to create more comprehensive representations. This multi-layer approach enables MVP-CBM to outperform state-of-the-art models like Explicd and other black-box methods across all evaluated datasets.

\begin{figure}[tbp]
  \centering
  \includegraphics[width=1.0\linewidth]{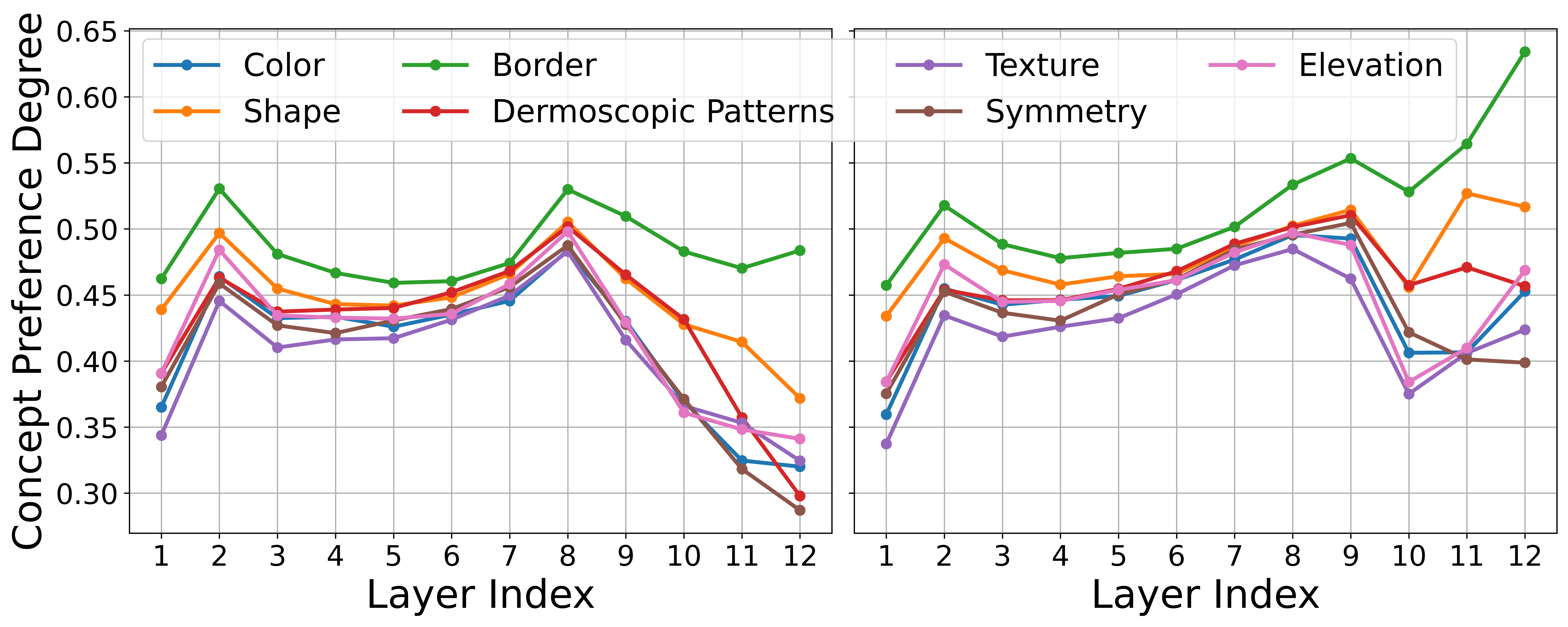}
  \caption{Examples of different concept preferences at each layer.}
  \label{figs:preference}
\end{figure}

\subsection{Ablation Study}
We conduct an ablation study on the ISIC2018 dataset to evaluate the effectiveness of various components and modules within the MVP-CBM, as shown in Table~\ref{tab:ablation_combined}. No. 2-4: Removing key components such as Intra-layer Concept Preference Modeling (ICPM), concept loss, and sparse loss led to significant declines in both BMAC and ACC, underscoring their essential roles in maintaining model performance.

\begin{figure}[tbp]
    \centering
    \includegraphics[width=\linewidth]{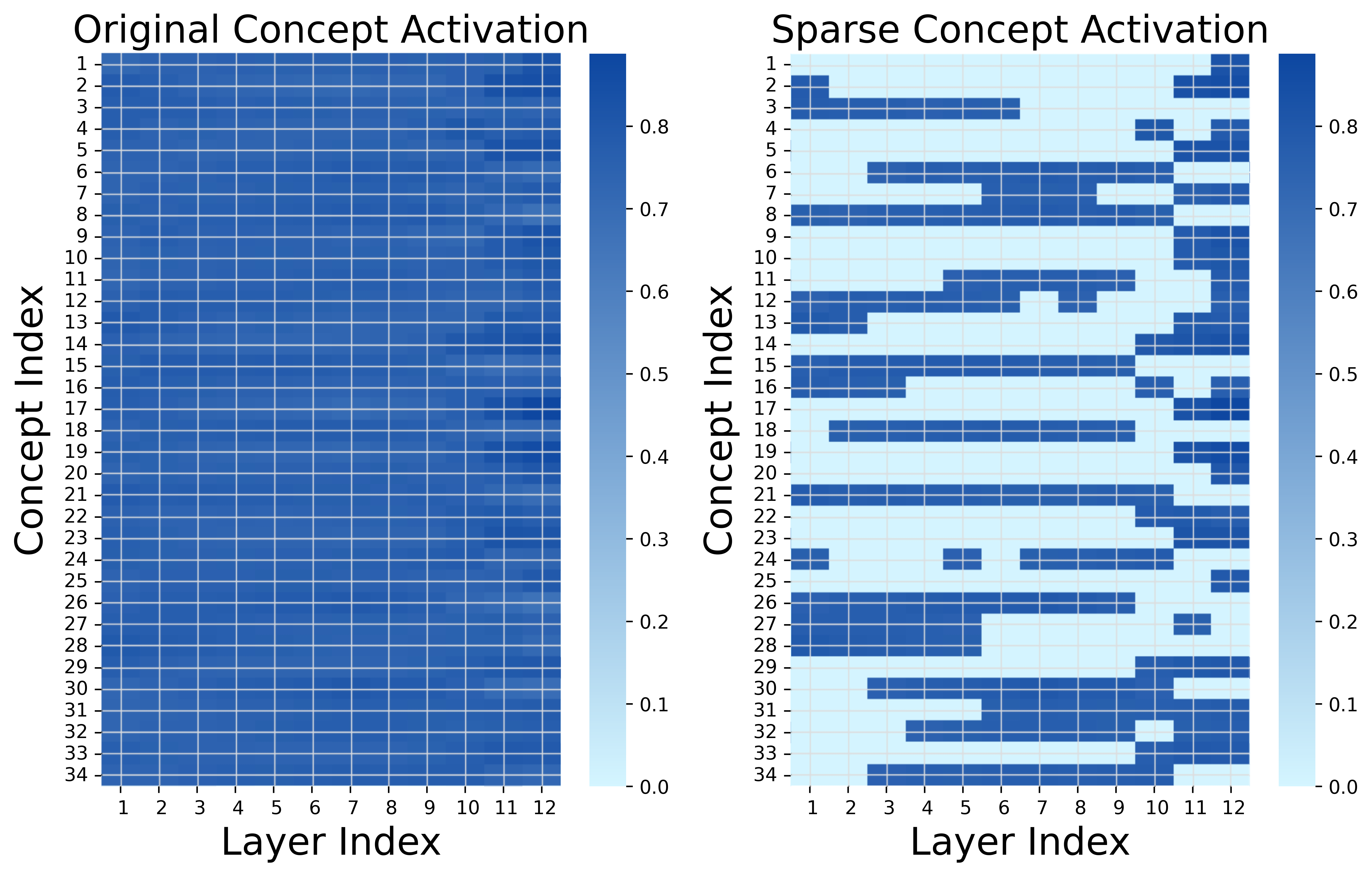}
    \caption{ The difference between original concept activation and our devised sparse concept activation.}
    \label{fig:sparse}
\end{figure}

For the impact of modifying specific modules. No.5: Replacing the adaptive pooling mechanism with a Q-former in the Multi-layer Concept Sparse Activation Fusion (MCSAF) results in a notable performance degradation, likely due to increased complexity and potential overfitting. Similarly, No.6: Substituting the hard mask with a soft mask in the MCSAF allows for more concepts to be activated simultaneously and introduces noise, diluting the focus on the most relevant concepts. No.7-8: Replacing sparse concept activation fusion with Layer Adaptive Weighted Fusion (LAWF) and a Multi-Head Attention Mechanism (MHAM) in the MCSAF further impairs performance. LAWF tends to overemphasize certain layers while neglecting others, leading to incomplete or imbalanced concept representations. MHAM fragments attention across different heads, hindering the coherent consolidation of concept activations.

For hyperparameter study. No.9-10: Removing the learnable parameters $\tau_1$ and $\tau_2$ from the ICPM and MCSAF results in a significant decline in model performance. In the ICPM, $\tau_1$ adjusts the importance of concept preferences, thereby enhancing the model’s ability to understand and differentiate high-level semantic concepts. Meanwhile, in the MCSAF, $\tau_2$ determines which concepts are activated in the multi-layer concept sparse activation fusion, thereby achieving sparsity.

\begin{table}[tbp]
    \centering
    \resizebox{\columnwidth}{!}{%
    \begin{tabular}{llcc}
        \toprule
        \textbf{No.} & \textbf{Model Variant} & \textbf{BMAC (\%)} & \textbf{ACC (\%)} \\
        \midrule
        \midrule
        1 & MVP-CBM (Ours) & \textbf{87.83±1.17} & \textbf{91.04±1.26} \\
        2 &   Ours w/o ICPM & 82.68±1.23 & 89.57±1.09 \\
        3 & Ours w/o $\mathcal{L}_{concept}$ & 78.83±0.91 & 84.84±0.95 \\
        4 & Ours w/o $\mathcal{L}_{sparse}$ & 70.21±0.76 & 82.76±0.94 \\
        
         \midrule
        \midrule
        5 & Q-former in MCSAF  & 85.02±0.92 & 89.17±0.83 \\
        6 & Soft Mask in MCSAF  & 83.29±0.82 & 89.27±0.85 \\
        7 & LAWF in MCSAF & 77.49±0.97 & 85.57±1.12 \\
        8 & MHAM in MCSAF & 63.37±0.76 & 73.54±0.82 \\
        \midrule
        \midrule
        9 & w/o $\tau_1$ in ICPM & 77.72±0.83 & 85.37±0.74 \\
        10 & w/o $\tau_2$ in MCSAF & 76.46±0.67 & 85.47±0.94 \\
        \bottomrule
    \end{tabular}
    }
    \caption{Ablation Study on ISIC2018, see Sec4.2 for abbreviation.}
     \label{tab:ablation_combined}
\end{table}

\subsection{Qualitative Analysis}
\noindent\textbf{Concept preference in ICPM.} 
Fig.~\ref{figs:preference} shows how concept preferences vary across layers of the model, highlighting the differential preference of various concepts at each layer. For example, in the left plot, the concept `Dermoscopic Patterns' exhibits the highest preference in the lower layers (Layer 8), whereas 
`Border' and `Shape' show lower preference across the early layers but tend to increase in higher layers (Layer 12). This indicates that certain concepts are more dominant at specific layers, supporting the idea that different concepts align better with different levels of the visual representation.
In the right plot, `Border' shows a relatively consistent preference across all layers, while concepts like `Texture' and `Symmetry' peak around Layer 8, indicating a stronger concept preference to intermediate layers. The varying preferences of these concepts across layers demonstrate that the final output layer does not necessarily contain the most optimal preference for all concepts.



\noindent\textbf{Sparse concept activation in MCSAF.}
Fig.~\ref{fig:sparse} compares the original concept activation and sparse concept activation across different layers of the model. On the left, the Original Concept Activation shows a dense activation pattern, where many concepts are activated across all layers, indicating a broad and less focused distribution of activations. In contrast, the Sparse Concept Activation on the right displays a more focused pattern, where only a few concepts are activated at each layer, leading to more concentrated and sparse activations. This difference highlights the effect of sparse activation, which reduces noise by activating only the most relevant concepts at each layer, thus improving interpretability by focusing on the most relevant information for predictions.

\begin{figure}[tbp]
  \centering
  \includegraphics[width=1.0\linewidth]{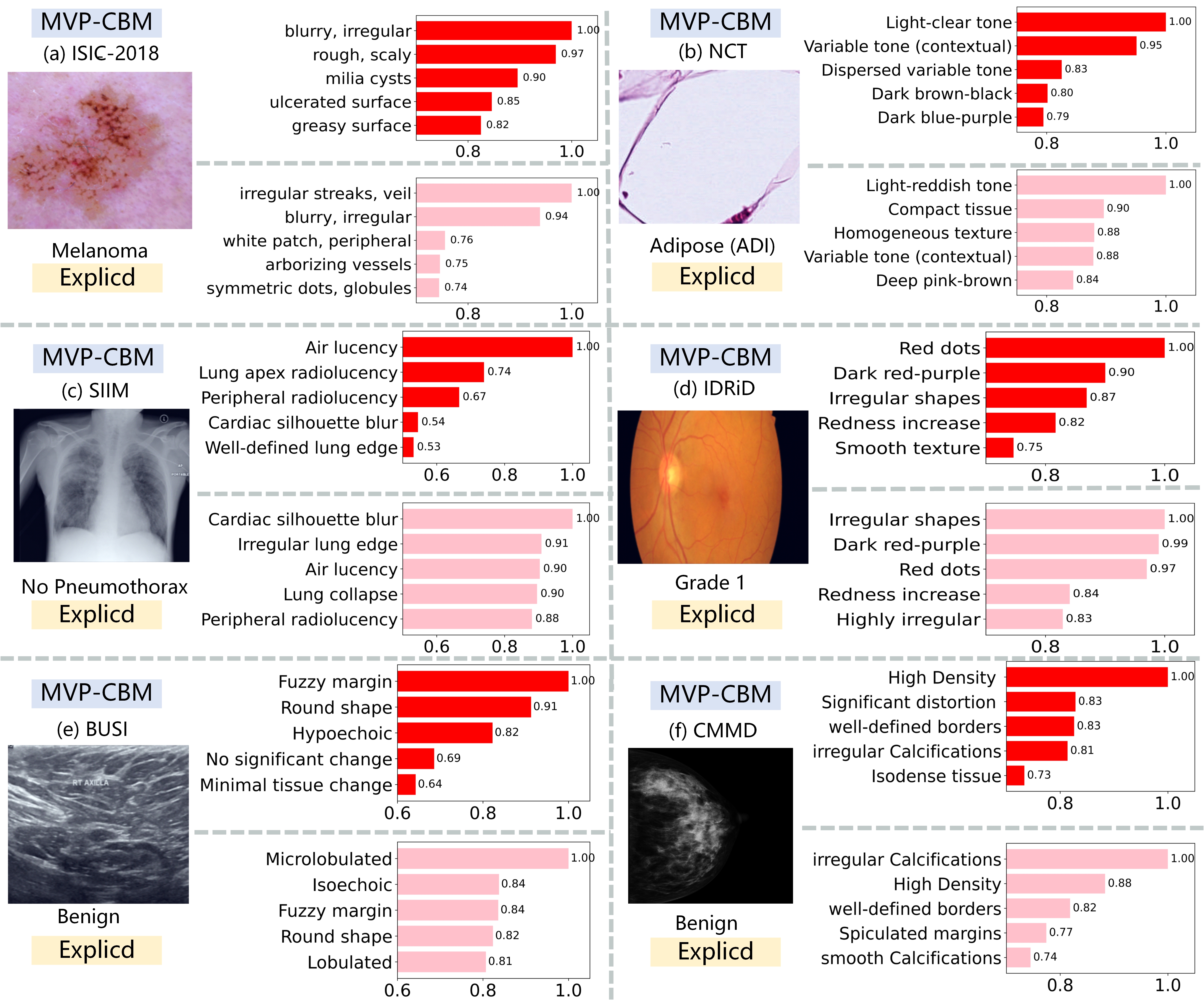}
  \caption{
 A comparison of our proposed MVP-CBM and the existing state-of-the-art Explicd generated interpretation with the top-5 contributed concepts.}
  \label{figs:vis}
\end{figure}

\subsection{Interpretability}

We visualize the concept bottleneck layer of both MVP-CBM and the state-of-the-art method Explicd~\cite{gao2024aligning}, focusing on how they rely on different concepts to make predictions. For each image across six datasets, we first normalize the activation scores of all concepts, then display the top 5 concepts with the highest activation scores in the concept bottleneck layer.

As shown in Fig.~\ref{figs:vis}-(f), MVP-CBM accurately diagnoses a Benign condition, as evidenced by the mammogram showing characteristics typical of benign masses such as `high density' and `well-defined borders'. High contribution values for concepts like `High Density' and `well-defined borders' further reinforce this diagnosis. In contrast, Explicd incorrectly labels the condition as Malignant, placing undue emphasis on `irregular Calcifications'. This misinterpretation overlooks the clear, smooth borders and absence of irregular calcifications that are hallmarks of benign lesions. Similar discrepancies between the two methods are observed across other datasets. Given that both the CM and SIIM datasets consist of chest X-ray images, we focus the visualization on the SIIM dataset. 

\section{Conclusions}
To the best of our knowledge, this is the first work to focus on the variance of concept preferences across visual layers within Concept Bottleneck Models (CBMs) to improve both model performance and interpretability. In this paper, we introduce the Multi-Layer Visual Preference Enhancement Concept Bottleneck Model (MVP-CBM), which incorporates two key components: Intra-layer Concept Preference Modeling (ICPM) and Multi-Layer Concept Sparse Activation Fusion (MCSAF). ICPM captures the varying preferences for concepts across different visual layers, allowing the model to leverage the most relevant layers for each concept. Meanwhile, MCSAF sparsely aggregates concept activations across layers, reducing noise and improving accuracy. Experimental results on multiple datasets demonstrate that our approach outperforms existing methods, setting a new standard for both model transparency and predictive performance.

\section*{Acknowledgments}
This work is supported by the National Natural Science Foundation of China under Grant 62271465, the China Postdoctoral Science Foundation under Grant 2024M763178, and the Suzhou Basic Research Program under Grant SYG202338.

\bibliographystyle{named}
\bibliography{ijcai25}

\end{document}